\documentclass[letterpaper]{article}
\usepackage{aaai16}
\usepackage{times}
\usepackage{helvet}
\usepackage{courier}
\usepackage{url}
\usepackage{graphicx}
\usepackage{hyperref}
\usepackage{amssymb}
\usepackage{amsmath} 
\usepackage{xcolor}
\hypersetup{
	colorlinks,
	linkcolor={red!50!black},
	citecolor={blue!50!black},
	urlcolor={blue!80!black}
}

\renewcommand{\phi}{\varphi}

\newcommand{\stam}[1]{}

\def\squarebox#1{\hbox to #1{\hfill\vbox to #1{\vfill}}}

\newcommand{\XORSample}{\ensuremath{\mathsf{XORSample }}}

\newcommand{\WeightMC}{\ensuremath{\mathsf{WeightMC}}}

\newcommand{\UniGen}{\ensuremath{\mathsf{UniGen}}}

\newcommand{\killthis}[1]{}
\newcommand{\prob}{\ensuremath{\mathsf{Pr}}}

\newcommand{\Relsat}{\ensuremath{\mathsf{Relsat}}}
\newcommand{\NP}{\ensuremath{\mathsf{NP}}}
\newcommand{\SAT}{\ensuremath{\mathsf{SAT}}}
\newcommand{\SMT}{\ensuremath{\mathsf{SMT}}}

\newcommand{\WISH}{\ensuremath{\mathsf{WISH}}}
\newcommand{\sharpSATTool}{\ensuremath{\mathsf{sharpSAT}}}
\newcommand{\Cachet}{\ensuremath{\mathsf{Cachet}}}
\newcommand{\ApproxCount}{\ensuremath{\mathsf{ApproxCount}}}

\newcommand{\SampleCount}{\ensuremath{\mathsf{SampleCount}}}
\newcommand{\BPCount}{\ensuremath{\mathsf{BPCount}}}
\newcommand{\MBound}{\ensuremath{\mathsf{MBound}}}
\newcommand{\CDP}{\ensuremath{\mathsf{CDP}}}
\newcommand{\HybridMBound}{\ensuremath{\mathsf{Hybrid}}-\ensuremath{\mathsf{MBound}}}

\newcommand{\MiniCount}{\ensuremath{\mathsf{MiniCount}}}
\newcommand{\sharpSAT}{\ensuremath{\#\mathsf{SAT}}}
\newcommand{\sharpP}{\ensuremath{\#\mathsf{P}}}
\newcommand{\approxMC}{\ensuremath{\mathsf{ApproxMC}}}
\newcommand{\ApproxMC}{\ensuremath{\mathsf{ApproxMC}}}

\newcommand{\SearchTreeSampler}{\ensuremath{\mathsf{SearchTreeSampler}}}
\newcommand{\SE}{\ensuremath{\mathsf{SE}}}
\newcommand{\SampleSearch}{\ensuremath{\mathsf{SampleSearch}}}

\newcommand{\US}{\ensuremath{\mathsf{US}}}
\newcommand{\SDD}{\ensuremath{\mathsf{SDD}}}

\newcommand{\CryptoMiniSAT}{\ensuremath{\mathsf{CryptoMiniSAT}}}
\newcommand{\satisfying}[1]{\ensuremath{R_{#1}}} %
\newcommand{\Vars}{\ensuremath{\mathsf{Vars }}}
\newcommand{\ProjectSatisfying}[2]{\ensuremath{{R_{#1}}_{|#2}}}

\makeatletter
\renewcommand\paragraph{\@startsection{paragraph}{4}{\z@}%
	{1.2ex \@plus1ex \@minus.2ex}%
	{-1em}%
	{\normalfont\normalsize\bfseries}}
\makeatother                                   
\newcommand{\ScalGen}{\ensuremath{\mathsf{UniGen2}}}
\newcommand{\loThresh}{\ensuremath{\mathrm{loThresh}}}
\newcommand{\hiThresh}{\ensuremath{\mathrm{hiThresh}}}
\newcommand{\MIS}{\ensuremath{\mathsf{MIS}}}
\newcommand{\WMC}{\ensuremath{\mathsf{WMC}}}
\newcommand{\UMC}{\ensuremath{\mathsf{UMC}}}
\newcommand{\CNF}{\ensuremath{\mathsf{CNF}}}
\begin{document}

\title{Constrained Sampling and Counting: Universal Hashing Meets SAT Solving\thanks{The order of authors is based on the number of published works contributed to the project and ties have been broken alphabetically.}}

\author{Kuldeep S. Meel$^1$ \And Moshe Y. Vardi$^1$ \And Supratik Chakraborty$^2$ \And  Daniel J. Fremont$^3$ \AND Sanjit A. Seshia$^3$   \And Dror Fried$^1$ \And Alexander Ivrii$^4$ \And Sharad Malik$^5$  
	\AND \\
	$^1$Department of Computer Science, Rice University
	$^2$Indian Institute of Technology, Bombay  \\    
	$^3$University of California, Berkeley 
	$^4$IBM Research, Haifa \\
	$^5$Princeton University
		 }
\maketitle
\begin{abstract}
Constrained sampling and counting are two fundamental problems in
artificial intelligence with a diverse range of applications, spanning
probabilistic reasoning and planning to constrained-random
verification. While the theory of these problems was thoroughly
investigated in the 1980s, prior work either did not scale to
industrial size instances or gave up correctness guarantees to achieve
scalability. Recently, we proposed a novel approach that combines
universal hashing and SAT solving and scales to formulas with hundreds
of thousands of variables without giving up correctness
guarantees. This paper provides an overview of the key ingredients of
the approach and discusses challenges that need to be overcome to
handle larger real-world instances.
\end{abstract}

\section{Introduction}

Constrained sampling and counting are two fundamental problems in
artificial intelligence. In constrained sampling, the task is to
sample randomly from the set of solutions of input constraints while the problem
of constrained counting is to count the number of solutions.  Both
problems have numerous applications, including in probabilistic
reasoning, machine learning, planning, statistical physics, inexact
computing, and constrained-random verification
\cite{Bacchus2003,JS96,NRJKVMS06,Roth1996}.  For example,
probabilistic inference over graphical models can be reduced to
constrained counting for propositional
formulas~\cite{Cooper90,Roth1996}. In addition, approximate
probabilistic reasoning relies heavily on sampling from
high-dimensional probabilistic spaces encoded as sets of
constraints~\cite{EGSS13a,JS96}.  Both constrained sampling and
counting can be viewed as aspects of one of the most fundamental
problems in artificial intelligence: exploring the structure of the
solution space of a set of constraints~\cite{RN09}.

Constrained sampling and counting are known to be computationally
hard~\cite{Valiant79,Jerr,Toda89}.  To bypass these hardness results,
approximate versions of the problems have been investigated.  Despite
strong theoretical and practical interest in approximation techniques
over the years, there is still an immense gap between theory and
practice in this area.  Theoretical algorithms offer guarantees on the
quality of approximation, but do not scale in practice, whereas
practical tools achieve scalability at the cost of offering weaker or
no guarantees.

Our recent work in constrained sampling and
counting~\cite{CFMSV14,CMV13a,CMV13b,CMV14,CFMSV14,CFMSV15,CFMV15,IMMV15},
has yielded significant progress in this area.  By combining the ideas
of using SAT solving as an oracle and the reduction of the solution
space via universal hashing, we have developed \emph{highly scalable}
algorithms that offer \emph{rigorous} approximation guarantees. Thus,
we were able to take the first step in bridging the gap between theory
and practice in approximate constrained sampling and counting.  A key
enabling factor has been the tremendous progress over the past
two decades in propositional satisfiability (SAT) solving, which makes
SAT solving usable as an algorithmic building block in practical
algorithms.

In this paper, we provide an overview of hashing-based sampling and counting techniques and put them in the context of related work. We then highlight key insights that have allowed us to further push the scalability envelope of these algorithms. Finally, we discuss challenges that still need to be overcome before hashing-based sampling and counting algorithms can be applied to large scale real-world instances. 

The rest of the paper is organized as follows. We introduce notation
and preliminaries in Section~\ref{sec:prelims} and discuss related
work in Section~\ref{sec:relatedWork}. We discuss key enabling
algorithmic techniques for the hashing-based approach in
Section~\ref{sec:enablingAlgorithmic}, followed by an overview of the
sampling and counting algorithms themselves in
Section~\ref{sec:hashing-sampling-counting}. We describe recent
advances in pushing forward the scalability of these algorithms in
Section~\ref{sec:scalability}, and finally conclude in
Section~\ref{sec:conclusion}.

\section{Preliminaries}\label{sec:prelims}
Let $F$ denote a Boolean formula in conjunctive normal form (CNF), and let
$X$ be the set of variables appearing in $F$.  The set $X$ is called
the \emph{support} of $F$.   We also use $\Vars(F)$ to denote the support of $F$.
Given a set of variables $S \subseteq X$ 
and an assignment $\sigma$ of truth values to the variables in $X$, 
we write $\sigma|_{S}$ for the projection of $\sigma$ onto $S$. 
A \emph{satisfying assignment} or \emph{witness} of $F$ is an assignment that 
makes $F$ evaluate to true.  
We denote the set of all witnesses of $F$ by 
$\satisfying{F}$ and the projection of $\satisfying{F}$ onto $S$ by 
{\ProjectSatisfying{F}{S}}. 
For notational convenience, whenever the formula $F$
is clear from the context, we omit mentioning it.

Let ${\mathcal I} \subseteq X$ be a subset of the support such that if two 
satisfying assignments $\sigma_1$ and $\sigma_2$ agree on ${\mathcal I}$, then $\sigma_1 = \sigma_2$. 
In other words, in every satisfying assignment, the truth values of variables in ${\mathcal I}$ 
uniquely determine the truth value of every variable in $X \setminus \mathcal{I}$. The
set ${\mathcal I}$ is called an \emph{independent support} of $F$, and 
$\mathcal{D} = X\setminus \mathcal{I}$ is a \emph{dependent support}. 
Note that there is a one-to-one correspondence between $\satisfying{F}$ and   
$\ProjectSatisfying{F}{\mathcal{I}}$.  There may be more than one independent support: 
$(a \vee \neg b) \wedge (\neg a \vee b)$ has three, namely $\{a\}$, $\{b\}$ and $\{a, b\}$. 
Clearly, if ${\mathcal{I}}$ is an independent support of $F$, so is every superset 
of ${\mathcal{I}}$. 

The \emph{constrained-sampling problem} is to sample randomly from $R_F$, given $F$.
A \emph{probabilistic generator} is a probabilistic algorithm that generates from
$F$ a random solution in $R_F$.  Let $\prob\left[E\right]$ denote the probability 
of an event $E$.  A \emph{uniform generator} $\mathcal{G}^{u}(\cdot)$ is a probabilistic 
generator that, given $F$, guarantees $\prob\left[\mathcal{G}^{u}(F) = y\right] = 1/|R_F|$,
for every $y \in R_F$.  An \emph{almost-uniform generator}
$\mathcal{G}^{au}(\cdot, \cdot)$ guarantees that for every $y \in R_F$,
we have $\frac{1}{(1+ \varepsilon)|R_F|}$ $\le$
$\prob\left[\mathcal{G}^{au}(F,\varepsilon) = y\right]\le$ $\frac{1 +
	\varepsilon}{|R_F|}$, where $\varepsilon > 0$ is a specified
\emph{tolerance}.  Probabilistic generators are allowed to occasionally ``fail" 
in the sense that no solution may be returned even if $R_F$ is non-empty.  
The failure probability for such generators must be bounded by a constant 
strictly less than 1.

The \emph{constrained-counting problem} is to compute the size of the
set $R_F$ for a given CNF formula $F$.  An  \emph{approximate counter} 
is a probabilistic algorithm ${\ApproxCount}(\cdot, \cdot, \cdot)$  that, 
given a formula $F$, a tolerance $\varepsilon>0$, and a confidence 
$1-\delta \in (0, 1]$, guarantees that 
$\prob\Big[|R_F|/(1+\varepsilon) \le {\ApproxCount}(F, \varepsilon,
1-\delta)\le (1+\varepsilon)|R_F|\Big] \ge 1-\delta$.

The type of hash functions used in the hashing-based approach to sampling and counting are 
\emph{$r$-universal} hash functions.
For positive integers $n$, $m$, and $r$, we write
$H(n,m,r)$ to denote a family of $r$-universal hash functions
mapping $\{0, 1\}^n$ to $\{0, 1\}^m$.  We use $h \xleftarrow{R}
H(n,m,r)$ to denote the probability space obtained by choosing a
hash function $h$ uniformly at random from $H(n,m,r)$.  The property
of $r$-universality guarantees that for all $\alpha_1, \ldots,
\alpha_r \in \{0,1\}^m $ and all distinct $y_1, \ldots, y_r \in
\{0,1\}^n$, $\prob\left[\bigwedge_{i=1}^r h(y_i) = \alpha_i\right.$
$\left.: h \xleftarrow{R} H(n, m, r)\right] = 2^{-mr}$.  
We use a particular class of such hash functions, denoted by $H_{xor}(n,m)$, which is 
defined as follows. 
Let $h(y)[i]$ denote the $i^{th}$ component of the vector $h(y)$.
This family of hash functions is then defined as
$\{h \mid h(y)[i] = a_{i,0} \oplus (\bigoplus_{k=1}^n a_{i,k}\cdot
y[k]), a_{i,k} \in \{0, 1\}, 1 \leq i \le m, \; 0 \leq k \leq n\}$, where
$\oplus$ denotes the XOR operation.  By choosing values of $a_{i,k}$
randomly and independently, we can effectively choose a random hash
function from $H_{xor}(n,m)$.  It was shown in~\cite{GSS07}
that this family is $3$-universal. 

\section{Related Work}\label{sec:relatedWork}
Constrained counting, the problem of counting the number of solutions of a 
propositional formula, is known as {\sharpSAT}. It is 
$\sharpP$-complete~\cite{Valiant79}, where {\sharpP} is 
the set of counting problems associated with {\NP} decision problems.
Theoretical investigations of {\sharpP} have led to the discovery of deep
connections in complexity theory
\cite{Toda89,Valiant79},
and there is strong evidence for its hardness \cite{AroBar09}.
It is also known that an efficient algorithm for constrained \emph{sampling}
would yield a fully polynomial randomized approximation scheme (FPRAS) 
for {\sharpP}-complete inference problems~\cite{JS96} -- a possibility 
that lacks any evidence so far and is widely disbelieved.  

In many applications of constrained counting, such as in probabilistic
reasoning, exact counting may not be critically important, and
approximate counts suffice.  Even when exact counts are important, the
inherent complexity of the problem may force one to work with
approximate counters.  Jerrum, Valiant, and Vazirani~\cite{Jerr} 
showed that approximate counting of solutions of CNF formulas,
to within a given tolerance factor, can be done with high confidence in
randomized polynomial time using an {\NP} oracle.  A key result
of~\cite{Jerr} states that for many problems, generating
solutions \emph{almost uniformly} is inter-reducible with approximate
counting; hence, they have similar complexity. Building on the Sipser's and Stockmeyer's early work~\cite{Sipser83,Stockmeyer83},  Bellare, Goldreich,
and Petrank~\cite{Bellare00} later showed that in fact, an {\NP}-oracle suffices for generating solutions of CNF formulas \emph{exactly uniformly} in randomized polynomial time.
 Unfortunately, 
these deep theoretical results have not been successfully reduced to
practice.  Our experience in implementing these techniques indicates
that they do not scale in practice even to small problem
instances involving few tens of variables~\cite{Meel14}.

Industrial approaches to constrained sampling in the context
of constrained-random verification~\cite{NRJKVMS06} either rely on Binary
Decision Diagram (BDD)-based techniques~\cite{Yuan2004}, which scale rather
poorly, or use heuristics that offer no guarantee of performance or 
uniformity when applied to large problem instances~\cite{KitKue2007}.
In prior academic works \cite{EGS12a,KGV83,GSS07,Selman-Sampling}, 
the focus is on heuristic techniques including Markov chain Monte Carlo (MCMC) methods and
techniques based on random seeding of {\SAT} solvers.
These methods scale to large problem instances, but either offer very weak or no guarantees on the uniformity 
of sampling, or require the user to provide hard-to-estimate problem-specific 
parameters that crucially affect the performance and uniformity of 
sampling~\cite{EGSS13b,EGSS13c,GD08,KitKue2007}.

The earliest approaches to {\sharpSAT} were based on DPLL-style {\SAT}
solvers and computed exact counts.  These approaches, e.g. {\CDP}~\cite{Birnbaum1999}, 
incrementally counted the number of solutions by introducing
appropriate multiplication factors for each partial solution found, eventually covering the entire solution space.
Subsequent  counters such as {\Relsat}~\cite{Bayardo97usingcsp},
{\Cachet}~\cite{Sang04combiningcomponent}, and
{\sharpSATTool}~\cite{Thurley2006} improved upon this by using
several optimizations such as component caching, clause learning, and the like.  Techniques based on BDDs and their
variants~\cite{minato93}, or d-DNNF
formulas~\cite{darwiche2004new}, have also been used to compute exact
 counts.  Although exact counters have been successfully
used in small- to medium-sized problems, scaling to larger problem
instances has posed significant challenges.  Consequently,
a large class of practical applications has remained beyond the reach
of exact counters~\cite{Meel14}.

To overcome the scalability challenge, more efficient techniques for
\emph{approximate} counting have been proposed.  The large majority of
approximate counters used in practice are \emph{bounding counters},
which provide lower or upper bounds but do not offer guarantees on the
tightness of these bounds. Examples include {\SampleCount}~\cite{gomes2007sampling},
{\BPCount}~\cite{KrocSabSel2008}, {\MBound} and
{\HybridMBound}~\cite{GSS06}, and {\MiniCount}~\cite{KrocSabSel2008}.
Another category of counters is called \emph{guarantee-less
  counters}. While these counters may be efficient, they provide no
guarantees and the computed estimates may differ from the exact
counts by several orders of magnitude~\cite{GHSS07}. Examples of
guarantee-less counters include {\ApproxCount}~\cite{wei2005new},
{\SearchTreeSampler}~\cite{ErGoSel2012}, {\SE}~\cite{Rubin2012}, and
{\SampleSearch}~\cite{GogDech2011}.

\section{Enabling Algorithmic Techniques}\label{sec:enablingAlgorithmic}

Hashing-based approaches to sampling and counting rely on three key algorithmic techniques -- one
classical, and two more recent: \emph{universal hashing},
\emph{satisfiability} (SAT) \emph{solving}, and
\emph{satisfiability modulo theories} ({\SMT}) \emph{solving}.

\paragraph{ Universal Hashing} Universal hashing is an algorithmic technique
that selects a hash function at random from a family of functions
with a certain mathematical property \cite{carter1977universal}. This technique
\emph{guarantees} a low expected number of collisions, even for an
arbitrary distribution of the data being hashed.  We have shown that universal hashing enables
us to partition the set $R_F$ of satisfying assignments of a formula
$F$ into roughly equally-sized ``small'' cells~\cite{CMV13a}. By choosing the definition
of ``small'' carefully, we can sample almost uniformly by first
choosing a random cell and then sampling uniformly inside the cell~\cite{CMV14}.
Measuring the size of sufficiently many randomly chosen cells also
gives us an approximation of the size of $R_F$~\cite{CMV13b}. To get a good
approximation of uniformity and count, we employ a \emph{3-universal}
family of hash functions.

\paragraph{SAT Solving} The paradigmatic {\NP}-complete problem of boolean 
satisfiability (SAT) solving \cite{Cook1971SAT}, is a central problem in
computer science. Efforts to develop practically successful SAT
solvers go back to the 1950s. The past 20 years have witnessed a
``SAT revolution" with the development of \emph{conflict-driven
  clause-learning} (CDCL) solvers \cite{BHMW09}.  Such solvers combine
a classical backtracking search with a rich set of effective
heuristics. While 20 years ago SAT solvers were able to solve
instances with at most a few hundred variables, modern SAT solvers solve
instances with up to \emph{millions} of variables in a
reasonable time \cite{MZ09}. Furthermore, SAT solving is continuing to 
demonstrate impressive progress \cite{Var14a}.  Propositional sampling 
and counting are both extensions of SAT; thus, practically effective 
SAT solving is a key enabler in our work.

\paragraph{SMT Solving} The Satisfiability Modulo Theories ({\SMT}) problem is
a decision problem for logical formulas in combinations of background
theories.  Examples of theories 
typically used  are the theory of real numbers, the theory of integers, 
and the theories of various data structures such as lists, arrays, bit vectors, 
and others.  {\SMT} solvers have shown dramatic progress over the past
couple of decades and are now routinely used in industrial software
development \cite{dMB11}.  Even though we focus on sampling and counting
for propositional formulas, we have to be able to combine propositional reasoning
with reasoning about hash values, which requires the power of {\SMT}
solvers. In our approach, we express hashing by means of XOR
constraints. These constraints can be reduced to CNF, but such a
reduction typically leads to computational inefficiencies during
SAT solving. Instead, we use {\CryptoMiniSAT}, an SMT solver which combines the
power of CDCL SAT solving with algebraic treatment of XOR
constraints, to yield a highly effective solver for a combination of
propositional CNF and XOR constraints \cite{SNC09}.
\section{Hashing-Based Sampling and Counting}\label{sec:hashing-sampling-counting}

In recent years, we have shown that the combination of
universal hashing and SAT/{\SMT} solving can yield a dramatic breakthrough in 
our ability to perform almost-uniform sampling and approximate counting
for industrial-scale formulas with \emph{hundreds of thousands}
of variables~\cite{CFMSV14,CMV13a,CMV13b,CMV14,CFMSV14,CFMSV15,CFMV15,IMMV15}.  The algorithms and
tools we developed provide the first scalable implementation with
provable approximation guarantees of these fundamental algorithmic 
building blocks. 

\paragraph{Approximate Counting} In \cite{CMV13b}, we introduced an
approximate constrained counter, called {\ApproxMC}.  The algorithm
employs XOR constraints to partition the solution space into ``small''
cells, where finding the right parameters to get the desired sizes of cells requires an
iterative search.  The algorithm then repeatedly invokes {\CryptoMiniSAT}  
to exactly measure the size of sufficiently many random cells (to achieve the desired confidence) 
and returns an estimate given by multiplying the median of the measured cell sizes by the number of cells covering the solution space.

\begin{figure}[h!b]
\centering
\includegraphics[scale=0.7]{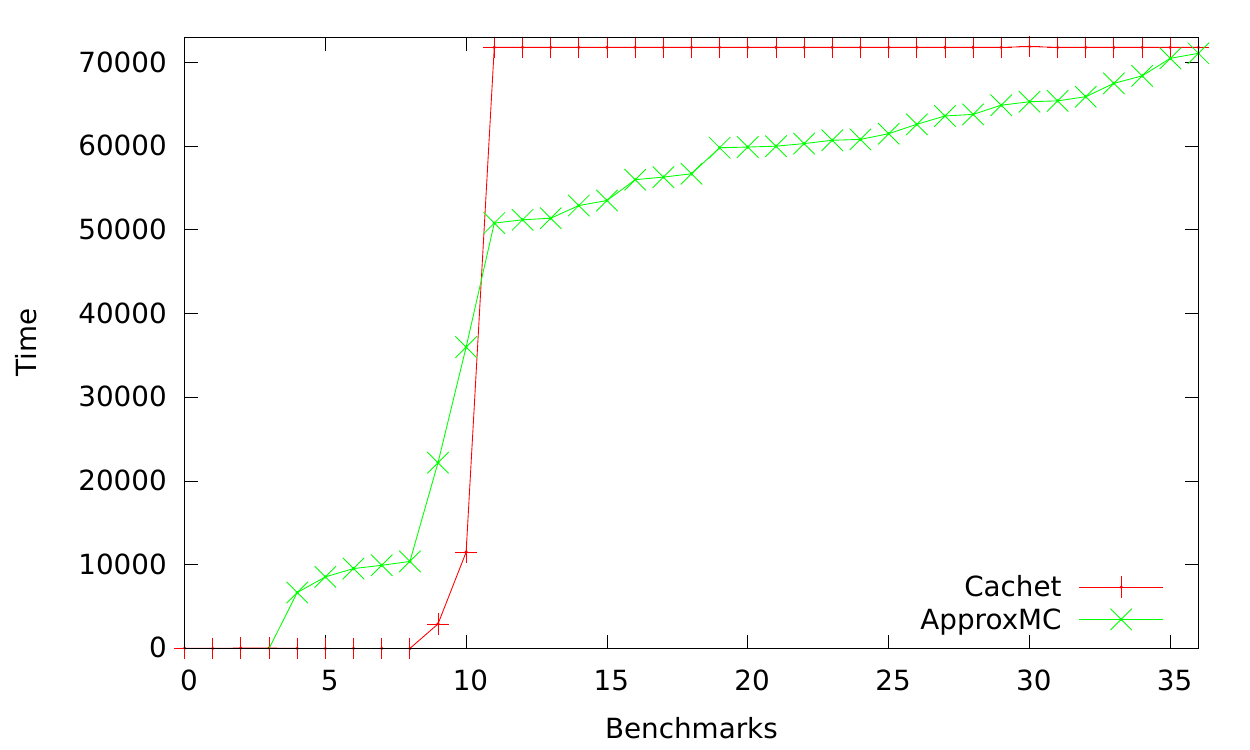}
\caption{Performance comparison between {\ApproxMC} and {\Cachet}.}
\label{approxmc-cachet}
\end{figure}

\begin{figure}[h!b]
	\centering
	\includegraphics[scale=0.7]{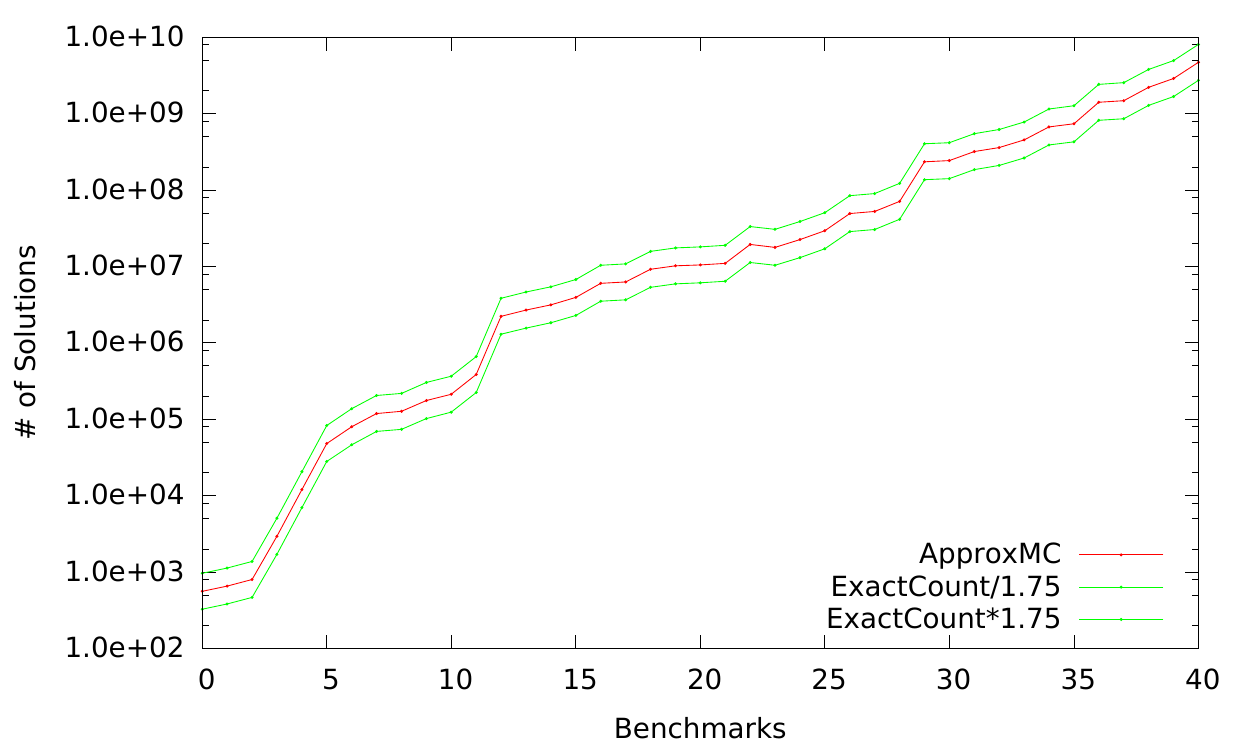}
	\caption{Quality of counts computed by {\approxMC}. The benchmarks are
		arranged in increasing order of model counts.}
	\label{fig:quality_comparison}
\end{figure}

\begin{figure}[h!b]
\centering
\includegraphics[scale=0.4]{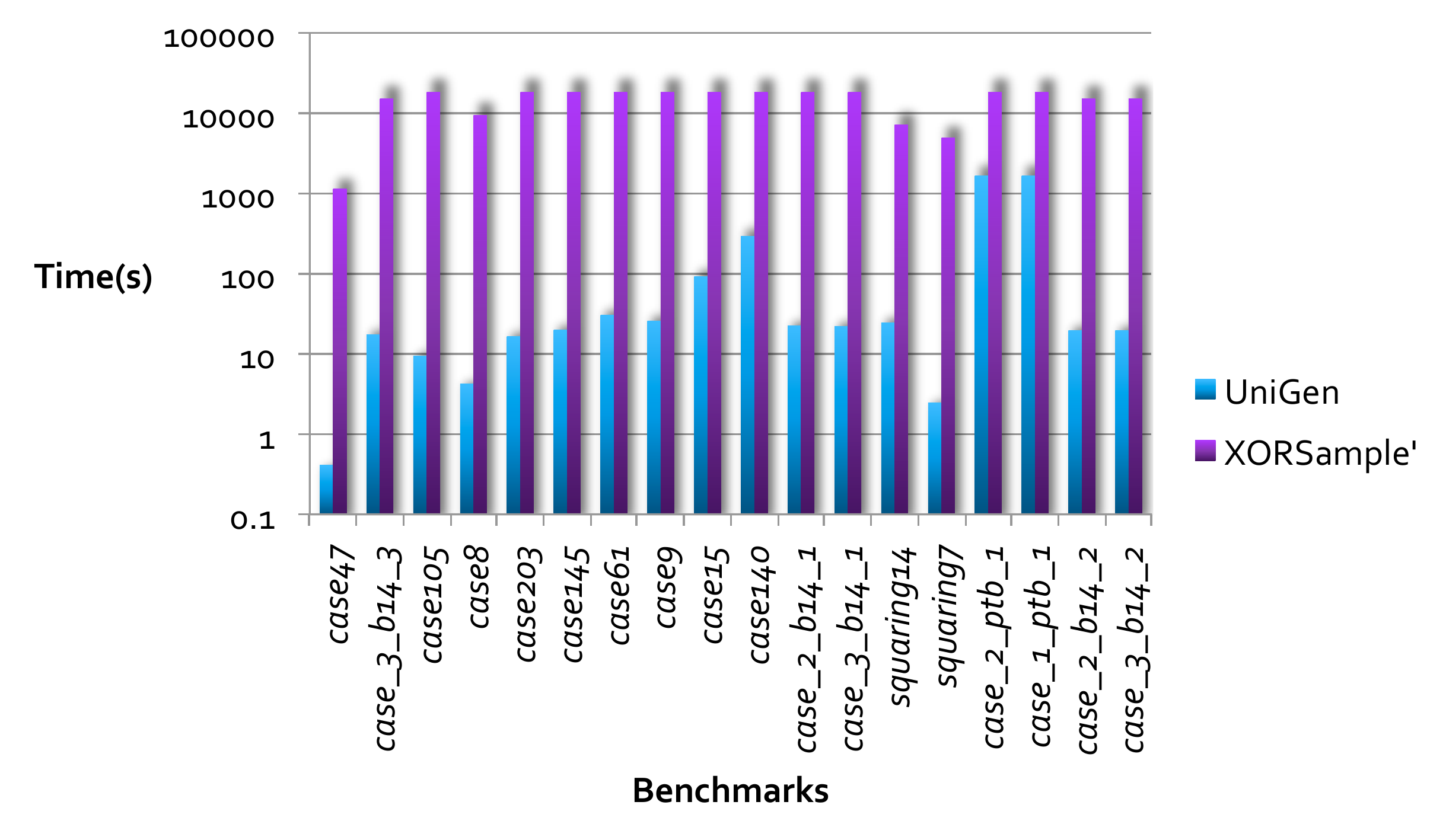}
\caption{Performance comparison between {\UniGen} and {\XORSample'}.} 
\label{unigen-xorsample}
\end{figure}

\begin{figure}[h!b]
	\centering
	\includegraphics[scale=0.6]{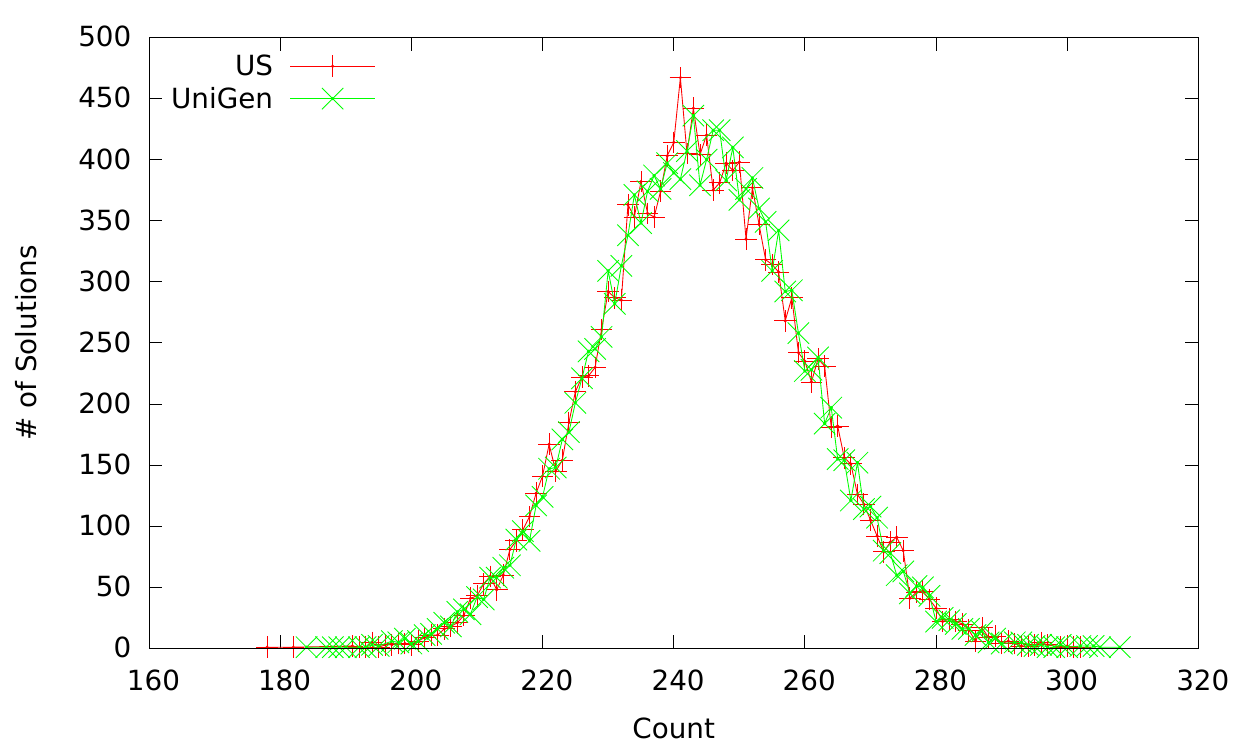}
	\caption{Histogram of solution frequencies comparing the distribution of {\UniGen} to that of an ideal uniform sampler (US).} 
	\label{uniformity}
\end{figure}

We compared {\ApproxMC} with {\Cachet}, a well known exact
counter~\cite{Sang04combiningcomponent}, using a timeout of 20 hours.
While {\Cachet} can solve  instances with up to 13K variables, 
{\ApproxMC} can solve very large instances with over 400K variables (see 
Fig~\ref{approxmc-cachet} for performance comparison over a subset of benchmarks). 
On instances where the exact count was available, the approximate count 
computed by {\ApproxMC} was within 4\% of the exact count, even when the tolerance requested was only 75\% (See Figure~\ref{fig:quality_comparison}).  
\paragraph{Almost-uniform Sampling} In \cite{CMV13a,CMV14}, we described an
almost-uniform propositional sampler called {\UniGen}.  The algorithm
first calls {\ApproxMC} to get an approximate count of the size of the
solution space. Using this count enables us to fine-tune the
parameters for using XOR constraints to partition the solution space
in such a way that a randomly chosen cell is expected to be small
enough so the solutions in it can be enumerated and sampled.
{\UniGen} was able to handle formulas with approximately 0.5M
variables. Its performance is several orders of magnitude better than
that of a competing algorithm called {\XORSample'}~\cite{GSS07}, which \emph{does not} offer an approximation
guarantee of almost uniformity (see Fig.~\ref{unigen-xorsample}).  To
evaluate the uniformity of sampling, we compared {\UniGen} with a uniform sampler (called {\US}, implemented by enumerating all solutions) on a benchmark with about 16K
solutions. We generated 4M samples, each from {\US} and {\UniGen};
the resulting distributions were statistically indistinguishable. (see Fig.~\ref{uniformity}).

\paragraph{Weighted Counting and Sampling} In \cite{CFMSV14}, we showed how
to extend the above algorithms for approximate counting and
almost-uniform sampling from the \emph{unweighted} case, in which all
assignments are given equal weight, to the \emph{weighted} case, in
which a weight function associates a weight with each assignment, and
counting and sampling must be done with respect to these weights.
Under some mild assumptions on the distribution of weights, we showed
that the unweighted algorithms for approximate counting and
almost-uniform sampling can be adapted to work in the weighted
setting, using only a SAT solver ({\NP}-oracle) and a black-box
weight function $w(\cdot)$.  For the algorithm to work well in
practice, we require that the \emph{tilt} of the weight function, which is
the ratio of the maximum weight of a satisfying assignment to the
minimum weight of a satisfying assignment, be small.  This is a
reasonable assumption for several important classes of problems~\cite{CFMSV14}. We were able to handle
formulas with over 60K variables, and our tools significantly
outperformed {\SDD}, a state-of-the-art exact weighted constrained counter
\cite{Dar11}.

\paragraph{Related recent work on hashing-based approaches}
The above work generated interest in the research community focused on
sampling and counting problems. We introduced the approach of
combining universal hashing with SAT solving for almost-uniform
sampling and approximate counting in~\cite{CMV13a,CMV13b}. Recently,
\cite{BVP15} proposed a probabilistic inference algorithm that is a
``simple reworking" (quoting authors) of {\WeightMC}.

 Building on our underlying approach, Ermon et
al.  suggested further improvements and proposed an algorithm, called {\WISH}, for
constrained  weighted
counting ~\cite{EGSS13b,EGSS13c}. The algorithm {\WISH} falls short of {\WeightMC} in terms of approximation guarantees and performance. Unlike {\WeightMC}, {\WISH} does not provide $(\varepsilon,\delta)$ guarantees and only provides constant-factor approximation guarantees. Furthermore, {\WISH} requires access to a most-probable-explanation 
(MPE) oracle, which is an optimization oracle and more expensive in practice than the {\NP} oracle 
(i.e. SAT solver) required by our approach~\cite{PD06}.
Recently, Zhu and Erom~\cite{ZE15} proposed an approximate algorithm, named RP-InfAlg, for approximate probabilistic inference.   This algorithm provides very weak approximation guarantees, and requires the use of hard-to-estimate parameters. Furthermore, their experiments were done with specific values of parameters, which are not easy to guess or compute efficiently.  The computational effort required in identifying the right values of the parameters is not addressed in their work. On the constrained sampling front, Ermon et al~\cite{EGSS13a} proposed the PAWS algorithm, which provides weaker guarantees than UniGen.  Like UniGen, PAWS requires a parameter estimation step.  However, given a propositional formula with n variables, UniGen uses $O(n)$ calls to an NP oracle to estimate the parameters, while PAWS requires $O(n.\log n)$ calls.

\section{Towards Scalable Sampling and Counting Algorithms}\label{sec:scalability}
We now discuss four recent promising directions towards further improving the scalability of hashing-based sampling and counting algorithms.

\paragraph{Parallelism}
There has been a strong recent revival of interest in parallelizing a wide variety of 
algorithms to achieve improved performance~\cite{Larus09}.  One of the main goals in 
parallel-algorithm design is to achieve a speedup nearly linear in the number of 
processors, which requires the avoidance of dependencies among different parts of the
algorithm~\cite{EZL89}.  Most of the sampling algorithms employed for  sampling fail 
to meet this criterion, and are hence not easily parallelizable.  
For example, by using constraint solvers with randomized branching heuristics, samples
with sequential dependencies are generated. Similarly, MCMC samplers often require a long sequential walk
before converging to the stationary distribution.  The lack of
techniques for sampling solutions of constraints in parallel while
preserving guarantees of effectiveness in finding bugs has been a 
major impediment to high-performance constrained-random verification (CRV). 

The algorithm {\ScalGen} presented in~\cite{CFMSV15} takes a step
forward in addressing this problem.  It has an initial preprocessing step that is
sequential but low-overhead,
followed by inherently parallelizable sampling steps.
It generates samples (stimuli) that are provably nearly as effective
as those generated by an almost-uniform sampler for purposes of detecting a
bug.  Furthermore, experimental evaluation over a diverse set of benchmarks demonstrates that the performance improvement for {\ScalGen} scales linearly with the number of processors: low communication overhead allows {\ScalGen} to achieve efficiency levels close to those of ideal distributed algorithms. Given that current practitioners are forced to trade
guarantees of effectiveness in bug hunting for scalability, the above
properties of {\ScalGen} are significant.  Specifically, they enable a
new paradigm for CRV wherein both stimulus generation and
simulation are done in parallel, providing the required runtime performance without sacrificing
theoretical guarantees.

\paragraph{Independence}
Another way to boost the performance of {\UniGen} is by reducing ``waste'' of  
hashed solutions. Currently, after {\UniGen} chooses a random cell of solutions, it selects
a single sample from that cell and throws away all the other solutions. The reason for
this ``wasting'' of solutions is to ensure that the samples obtained are not only
almost-uniformly distributed, but also \emph{independently} distributed. Independence is an 
important feature of sampling, which is required to ensure that probabilities are multiplicative. 
(For example, in a sequence of Bernoulli coin tosses, independence is required to ensure that an 
outcome of heads is almost-certain in the limit.) In MCMC sampling, ensuring independence requires
very long random walks between samples.  An important feature of {\UniGen} is that its 
almost-uniform samples are independent. Such independence, however, comes at the cost
of throwing away many solutions. 

In some applications, such as constrained-random verification, full independence between successive samples is not needed: it is only required that the samples provide good coverage over the space of all solutions. In~\cite{CFMSV15}, our proposed algorithm, {\ScalGen}, gains efficiency by relaxing independence, while still maintaining provable coverage guarantees.
Specifically, a cell is ``small'' in the sense discussed earlier when the number of solutions it contains is between parameters {\loThresh} and {\hiThresh}, both computed from the tolerance $\varepsilon$.
Instead of picking only one solution from a small cell, {\ScalGen} randomly picks {\loThresh} distinct solutions.
This leads to a theoretical guarantee that {\ScalGen} requires significantly fewer SAT calls than {\UniGen} to obtain a given level of bug-finding effectiveness in constrained-random verification.

\paragraph{Independent Support}
These hashing-based techniques crucially rely on the ability of combinatorial solvers 
to solve propositional formulas represented as a conjunction of the input formula and 
3-universal hash functions.
Due to our use of $H_{xor}$, this translates to \emph{hybrid}
formulas which are conjunctions of CNF and XOR clauses. Therefore, a key challenge  is to further speed up search for the solutions of
such hybrid formulas.  The XOR clauses invoked here are \emph{high-density} clauses, 
consisting typically of $n/2$ variables, where $n$ is the number of variables in the input formula.
Experience has shown that the high density of the XOR clauses plays a major role
in runtime performance of  solving hybrid formulas.

Recently, Ermon et al. introduced a technique to reduce the density of XOR constraints.  Their technique gives hash functions with properties that are weaker than 2--universality, but  sufficient for approximate counting~\cite{EGSS14a}.  In practice, however, this approach does not appear to achieve significant reduction in the density of XOR constraints and, therefore, the practical impact on runtime is not significant. Furthermore, It is not yet clear whether the constraints resulting from Ermon et al's techniques~\cite{EGSS14a} can be used for constrained sampling, while providing the same theoretical guarantees that our work gives.

Motivated by the problem of high-density XOR constraints, we proposed the idea of restricting the hash functions to only the independent support $\mathcal{I}$ of a formula, and showed that even with this restriction, we obtain the required universality properties of hash functions needed for constrained sampling and counting techniques~\cite{CMV14}.
Since many practical instances admit independent supports much smaller than the total number of variables (e.g. we may drop all variables introduced by Tseitin encoding), this often allows the use of substantially less dense XOR constraints.
While it is possible in many cases
to obtain an over-approximation of $\mathcal{I}$ by examining the domain from which the instance 
is derived, the work in~\cite{CMV14} does not provide an algorithmic approach for determining $\mathcal{I}$,
and experience has shown that the manual approach is error prone.

In~\cite{IMMV15}, we proposed the first algorithm, called {\MIS}, to find minimal independent supports. The key idea of this algorithmic procedure is the reduction of the problem of minimizing an independent support of a Boolean formula to Group Minimal Unsatisfiable Subset (GMUS). To illustrate the practical value of this approach, we used {\MIS} to compute a minimal independent support for each of our {\ScalGen} and {\ApproxMC} benchmarks, and ran both algorithms using hashing only over the computed supports. 
Over a wide suite of benchmarks, experiments demonstrated that this 
hashing scheme improves the runtime performance of {\ScalGen} and {\ApproxMC} by two to three orders of magnitude. It is worth noting that these runtime improvements come at no cost of theoretical guarantees:
both {\ScalGen} and {\ApproxMC} still provide the same strong theoretical guarantees.

\paragraph{From Weighted to Unweighted Model Counting}
Recent approaches to weighted model counting ({\WMC}) have focused on adapting unweighted model counting ({\UMC}) techniques
to work in the weighted
setting~\cite{SangBearKautz2005,XCD12,CFMSV14}.  
Such adaptation requires intimate understanding of the implementation details of the {\UMC}
techniques, and on-going maintenance, since some of these techniques evolve over time.
In~\cite{CFMV15}, we flip this approach and present an efficient reduction of
literal-weighted {\WMC} to {\UMC}. The reduction preserves the normal
form of the input formula, i.e. it provides the {\UMC} formula in the
same normal form as the input {\WMC} formula.  
Therefore, an important contribution of our reduction is to provide a {\WMC}-to-{\UMC} module that
allows any {\UMC} solver, viewed as a \emph{black box}, to be
converted to a {\WMC} solver. This enables the automatic leveraging of
progress in {\UMC} solving to make progress in {\WMC} solving.

We have implemented our {\WMC}-to-{\UMC} module on top of
state-of-the-art exact unweighted model counters to obtain exact
weighted model counters for {\CNF} formulas with literal-weighted
representation.  Experiments on a suite of benchmarks indicate that
the resulting counters scale to significantly larger problem instances
than what can be handled by a state-of-the-art exact weighted model
counter~\cite{CD13}.  Our results suggest that we can
leverage powerful techniques developed for SAT  and related domains in recent years to handle
probabilistic inference queries for graphical models encoded as {\WMC}
instances. 
Furthermore, we
demonstrate that our techniques can be extended to more general
representations where weights are associated with constraints instead
of individual literals. 

\section{Conclusion}\label{sec:conclusion}
Constrained sampling and counting problems have a wide range of applications in artificial intelligence, verification, machine learning and the like. In contrast to prior work that failed to provide scalable algorithms with strong theoretical guarantees, hashing-based techniques offer scalable sampling and counting algorithms with rigorous guarantees on quality of approximation. Yet, many challenges remain in making this approach more viable for real-world 
problem instances. Promising directions of future research include designing efficient hash functions and developing SAT/SMT solvers specialized to handle constraints arising from these techniques. 
\fontsize{9.5pt}{10.5pt} \selectfont
\end{document}